\documentclass[10pt,twocolumn,letterpaper]{article}

\usepackage{iccv}
\usepackage{times}
\usepackage{epsfig}
\usepackage{graphicx}
\usepackage{amsmath}
\usepackage{amssymb}

\usepackage{xspace} 
\usepackage{color}
\usepackage{bm}

\usepackage{subcaption}
\usepackage{multirow}
\usepackage{cases}
\usepackage[ruled]{algorithm2e}
\usepackage{booktabs} 
\SetKwInOut{Input}{Input}
\SetKwInOut{Output}{Output}
\DontPrintSemicolon

\usepackage[export]{adjustbox}
\usepackage{diagbox}
\usepackage{pgf}
\usepackage[accsupp]{axessibility}  
\usepackage{stackengine}
\usepackage[pagebackref=true,breaklinks=true,letterpaper=true,colorlinks,bookmarks=false]{hyperref}

\usepackage[disable]{todonotes}  

\iccvfinalcopy 

\def\iccvPaperID{7231} 

\ificcvfinal\fi

\def\name{{HollowNeRF}\xspace}
\begin{document}

\title{HollowNeRF: Pruning Hashgrid-Based NeRFs\\ with Trainable Collision Mitigation}

\author{Xiufeng Xie, Riccardo Gherardi, Zhihong Pan, Stephen Huang\\
Oppo Mobile Telecommunications Corp.\\
2479 E Bayshore Rd, Palo Alto, CA, 94303, USA\\
{\tt\small xxie28@gmail.com}
}

\maketitle
\ificcvfinal\thispagestyle{empty}\fi

\begin{abstract}
\noindent Neural radiance fields (NeRF) have garnered significant attention, with recent works such as Instant-NGP accelerating NeRF training and evaluation through a combination of hashgrid-based positional encoding and neural networks. However, effectively leveraging the spatial sparsity of 3D scenes remains a challenge. To cull away unnecessary regions of the feature grid, existing solutions rely on prior knowledge of object shape or periodically estimate object shape during training by repeated model evaluations, which are costly and wasteful. 
To address this issue, we propose HollowNeRF, a novel compression solution for hashgrid-based NeRF which automatically sparsifies the feature grid during the training phase. Instead of directly compressing dense features, HollowNeRF trains a coarse 3D saliency mask that guides efficient feature pruning, and employs an alternating direction method of multipliers (ADMM) pruner to sparsify the 3D saliency mask during training. By exploiting the sparsity in the 3D scene to redistribute hash collisions, HollowNeRF improves rendering quality while using a fraction of the parameters of comparable state-of-the-art solutions, leading to a better cost-accuracy trade-off. Our method delivers comparable rendering quality to Instant-NGP, while utilizing just 31\% of the parameters. In addition, our solution can achieve a PSNR accuracy gain of up to 1dB using only 56\% of the parameters.
\end{abstract}

\section{Introduction}
\label{sec:intro}

\begin{figure}[t]
\begin{center}
\includegraphics[width=0.99\linewidth]{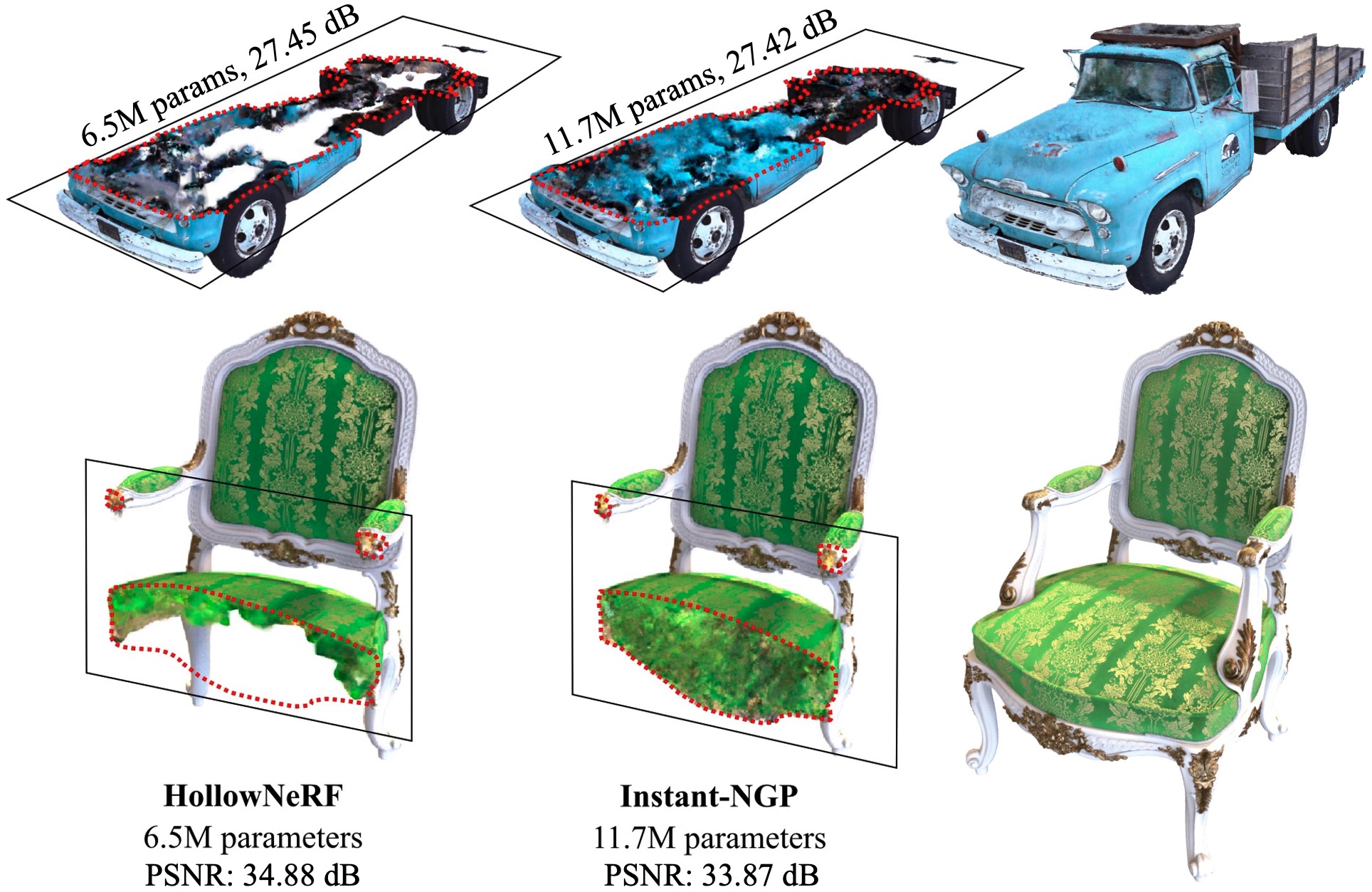}
\end{center}
   \caption{Cross-section views of rendered 3D scenes. The left column shows the hollow interior of \name's rendering, while the middle column shows the solid interior of Instant-NGP's rendering. The right column displays \name's rendering without slicing.}
\label{fig:hollow_chair}
\end{figure}

\noindent Neural Radiance Fields (NeRF~\cite{mipnerf, nerf}) have gained widespread recognition across academia and industry due to their remarkable capability to generate photorealistic novel views of 3D scenes from a collection of 2D images. Inspired by the volumetric representation, NeRF models the scene as a continuous 5D plenoptic function, enabling the creation of high-fidelity renderings with accurate lighting and shading effects. This technique has found versatile applications in fields such as computer graphics, virtual and augmented reality, as well as robotics. 

Training and evaluating NeRF models can be computationally expensive, and many recent works~\cite{kilo, ngp, tensorf, plenoxels, donerf, dvgo} have focused on improving the efficiency of NeRF.
Instant-NGP~\cite{ngp} is an established solution with state-of-the-art training speed. It employs a lightweight hashgrid for input encoding and a small multi-layer perceptron (MLP) to disambiguate hash collisions. However, the lightweight hashgrid unavoidably causes severe collisions at fine resolutions. These collisions are evenly scattered across the occupied voxels, resulting in a suboptimal accuracy. 
Another line of research focuses on accelerating NeRF rendering by only sampling near the surface of interest, but has to rely on prior knowledge of surface geometries (either from conventional algorithms such as shape-from-silhouette~\cite{convnerf} or an MLP predicting the depth distribution along each camera ray~\cite{donerf}).
However, a coarse surface estimation degrades NeRF rendering quality, while a precise surface estimation adds too much complexity, defeating the purpose of acceleration.

We propose \name, a novel NeRF compression method using trainable hash collision mitigation to improve rendering accuracy while consuming less parameters than existing NeRF methods. 
Built on a hash-based pipeline from Instant-NGP, \name prioritizes the important features (of visible voxels) and prunes unnecessary features (of invisible voxels), leading to a redistribution of hash collision probability across the 3D volume. 
When two voxels direct to the same hash bucket, Instant-NGP shapes the shared feature in this bucket as a mixture of the desired features, which harms the accuracy. In contrast, \name steers the shared feature to fit the more important voxel, and prunes the feature of the less important voxel toward $\bm{0}$, reducing interference to features sharing the same bucket. 
Specifically, when reading the feature of a certain voxel, \name further scales the feature by a trainable \emph{saliency weight} whose value captures the voxel's visibility. To reduce the cost, we divide the 3D space into coarse grid regions and assign a saliency weight to capture each region's visibility, forming a trainable 3D saliency grid.
Unlike existing methods~\cite{convnerf, donerf} that require prior knowledge of the surface geometries, \name learns to prioritize the important features by training the saliency grid, which converges to a ``hollow'' saliency distribution across the 3D volume.
Figure~\ref{fig:hollow_chair} showcases this ``hollow'' rendering result.
 
The proposed design consists of three main components: a trainable 3D saliency grid to guide the compression of dense features (\S\ref{sec:mask}); a soft zero-skipping gate that enhance the MLP in the NeRF model to ensure a feature compressed to $\bm{0}$ translates to a zero density in the 3D space (\S\ref{sec:gate}); a pruner to further push unnecessary features to exact $\bm{0}$ instead of a small non-zero value by alternating direction method of multipliers or ADMM~\cite{admm0} (\S\ref{sec:admm}). Our experiments demonstrate that \name achieves better accuracy (PSNR and LPIPS) than state-of-the-art methods while using significantly fewer parameters.  

The key contributions of this work are:
\begin{itemize}
\item We propose a novel NeRF compression solution, \name, that learns to prioritize features defining the visible surface and prune invisible internal features without prior knowledge of surface geometries.

\item We use an ADMM-based optimization framework to prune unnecessary features during NeRF training and enhance the MLP with a soft zero-skipping gate to ensure that pruned features correctly map to zero density.

\item  We evaluate the performance of \name on popular NeRF datasets and our solution demonstrates a significantly superior balance between cost and accuracy than state-of-the-art solutions.
\end{itemize}

\section{Background and related work}
\label{sec:related}
\begin{figure*}[t]
\begin{center}
\includegraphics[width=0.90\linewidth]{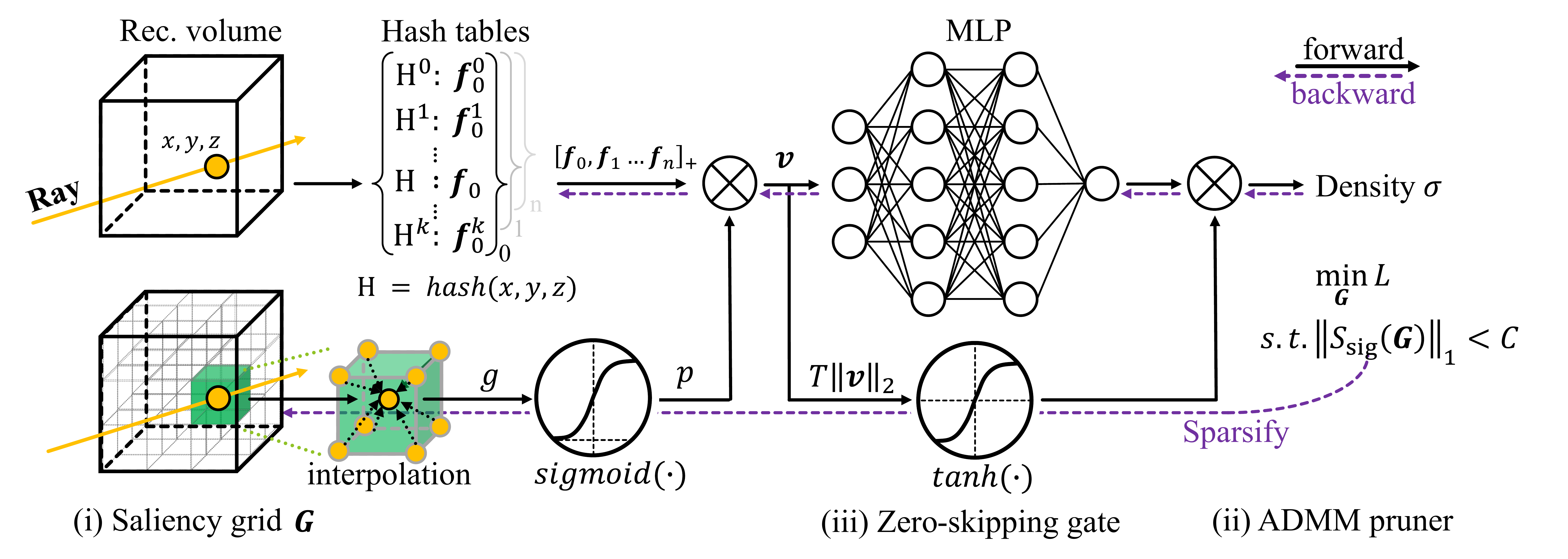}
\end{center}
   \vspace{-1.5em}
   \caption{Overview of the \name workflow.
}
\label{fig:design_all}
\end{figure*}

\noindent Differentiable rendering ~\cite{nerf, diffstereo, volumetricBA, plenoxels, nvdiffrec} has emerged as a prominent alternative framework for novel view synthesis, alongside the conventional 3D rendering approaches. 
Following the seminal NeRF paper~\cite{nerf}, the research community has matured the approach and enables it to handle arbitrarily large scenes~\cite{blocknerf, bungeenerf, meganerf, mipnerf360}, complex reflections~\cite{reflective, netrf, nerfosr, refnerf, nerfinthedark},and small training datasets with a limited number of views~\cite{dietnerf, depthnerf, regnerf}.
Recent works also substantially improve the efficiency of NeRF, reducing the run-time complexity at training~\cite{ngp, plenoxels} and inference~\cite{ngp, adanerf, donerf, pointnerf, hrsrg, plenoctrees, baking}. 

Comparatively fewer works have tackled the problem of maximizing visual fidelity for a given space complexity budget~\cite{ngp, torch-ngp1}, which is the focus of this paper.
The original NeRF technique~\cite{nerf} encodes the entire 3D scene in the weights of an MLP that predicts opacity ($\sigma$) and color $(r, g, b)$ given a 3D position ($x$, $y$, $z$) and direction ($\theta$, $\phi$). While spatial complexity ( $438$K parameters) is not the primary bottleneck, the substantial limitation of NeRF lies in the remarkably slow training and inference speeds. This is primarily attributed to the deep MLP with 8 hidden layers, each having a width of 256 neurons.  
It has since been shown~\cite{plenoxels} that differentiable rendering does not necessarily require neural networks. In~\cite{plenoxels}, a scene is encoded as a neuron-free sparse voxel grid of opacities and RGB spherical harmonics. Without the MLP, training and inference run fast, at the cost of a significantly high spatial complexity.

In the continuum between the two extremes discussed above, there are techniques that harness the strengths of both ends and fare better in the balance between training speed and spatial complexity.
Instant-NGP~\cite{ngp} is a representative example, which encodes features using a multi-resolution hashgrid and then feed the features into a lightweight MLP with only 2 hidden layers to decode the color and density; both the features in the hashgrid and the MLP are trained in conjunction.
Our work extends the hash-based pipeline from Instant-NGP by achieving higher quality (PSNR \& LPIPS) with less number of parameters.
We achieve this goal by reclaiming the resources spent on empty, invisible, or internal regions. Instead of using an auxiliary MLP to predict space occupancy like DONeRF~\cite{donerf}, we learn it through a trainable lightweight volumetric saliency grid.

Some existing literature has applied model compression techniques to NeRF: CC-NeRF~\cite{torch-ngp1} uses tensor decomposition to obtain a low-rank approximation of the learned network. Similarly, TensoRF~\cite{tensorf} represents the volume as a 4D tensor and factorizes it into low-rank components. Unlike these compression works, \name vets information based on its actual impact on the rendering accuracy, rather than raw entropy. The relative performance of these approaches are investigated in \S\ref{sec:vsngp}.

\section{\name design}
\label{sec:design}

\begin{figure}[t]
\begin{subfigure}[b]{0.48\linewidth}
    \centering
    \includegraphics[width=\linewidth]{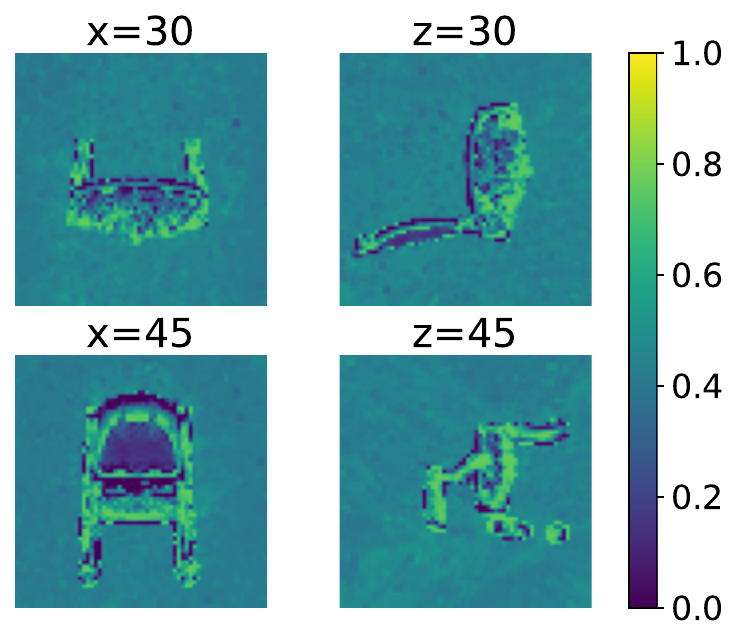}
    \caption{Without ADMM pruner.}
\label{fig:slice_no_pruner}
  \end{subfigure}
  \begin{subfigure}[b]{0.48\linewidth}
  \centering
    \includegraphics[width=\linewidth]{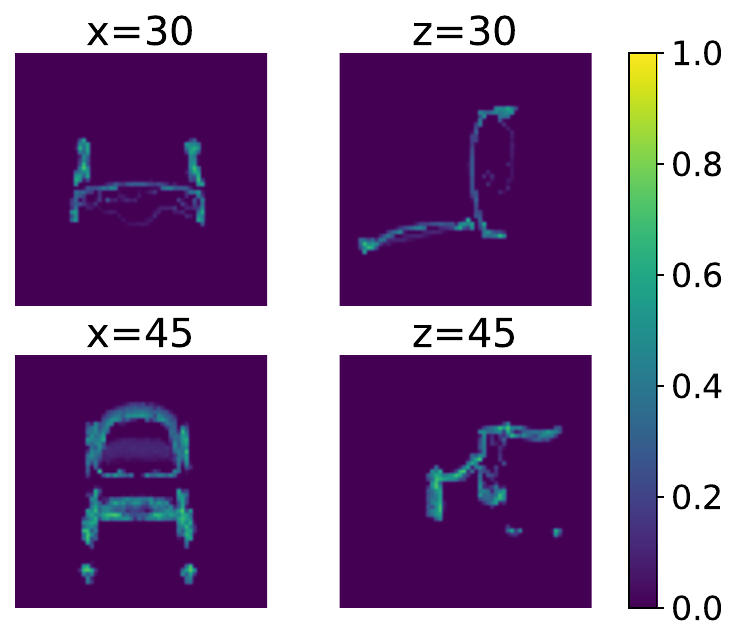}
    \caption{With ADMM pruner.}
\label{fig:slice_with_pruner}
  \end{subfigure}
   \caption{Slices of a $64\times64\times64$  saliency grid trained on the ``Chair'' scene from the NeRF synthetic dataset\cite{nerf}.}
\label{fig:grid_slice}
\end{figure}

\noindent This section presents the system design of \name, as outlined in Figure~\ref{fig:design_all}: \emph{(i)} Given an input coordinate $\bm{x}=(x, y, z)$, we fetch the corresponding feature $\bm{f}$ from the multi-level hashgrid, following the Instant-NGP method~\cite{ngp}.
\emph{(ii)} We predict the saliency weight $p$ of position $\bm{x}$  by processing the information from a trainable 3D saliency grid $\mathcal{G}$ through trilinear interpolation and a sigmoid function  (\S\ref{sec:mask}), and use $p$ to scale the feature $\bm{f}$ by $\bm{v}=p\bm{f}$.
Figure~\ref{fig:slice_no_pruner} showcases a trained saliency grid $\mathcal{G}$, demonstrating that training $\mathcal{G}$ can mitigate hash collisions by suppressing the unnecessary features in empty or invisible regions.
\emph{(iii)} The weighted feature $\bm{v}$ is fed into an MLP, which decodes $\bm{v}$ to obtain the density $\sigma$ and color at position $\bm{x}$. To ensure that $\bm{v}=\bm{0}$ maps to $\sigma=0$, we introduce a zero-skipping gate (\S\ref{sec:gate}).
\emph{(iv)} The density and color outputs of the MLP are used for volume rendering, and the resulting image is compared to the ground truth to obtain a loss function $L$. During training, an ADMM pruner (\S\ref{sec:admm}) sparsifies the saliency value distribution across the 3D space by enforcing a sparsity constraint when optimizing the loss $L$.
Figure~\ref{fig:slice_no_pruner} shows that a large portion of the unnecessary features are not entirely eliminated by training the saliency grid, with saliency weights diminished but not reaching $0$.
Therefore, we introduce the ADMM pruner to explicitly prune saliency weights of the empty or invisible regions to $0$, as shown in Figure~\ref{fig:slice_with_pruner}.

\subsection{Trainable 3D saliency grid}
\label{sec:mask}
\noindent In typical 3D scenes without large transparent objects, most regions of the space do not contribute to the final rendering as they are either empty or invisible from any view angle. To leverage this sparsity, methods such as Instant-NGP use a coarse-grained binary mask to capture space occupancy, which is periodically updated during training by evaluating the NeRF model at each voxel and checking if the density is above a threshold. This occupancy mask guides sampling in ray marching by skipping unoccupied regions. However, it still keeps unnecessary features in internal occluded regions that have non-zero densities but make no contributions to the visible surfaces from any view angle, as shown in Figure~\ref{fig:hollow_chair}. 
 
With this insight, we propose to reclaim the capacity spent on internal regions by learning the sparsity pattern through training, effectively making objects \emph{hollow}.
The benefits of this approach are twofold: it reduces space complexity by requiring fewer features, and improves accuracy by reducing hash collisions.
However, precisely locating the invisible regions of objects can be challenging without any prior knowledge of the object's shape.

To addresses this challenge, our approach utilizes the hashgrid-based positional encoding from Instant-NGP~\cite{ngp} and extends it by introducing a trainable saliency grid $\mathcal{G}$ with a resolution of $T\times T\times T$. Specifically, \name trains a 3D tensor data structure $\mathcal{G}$ alongside the NeRF model (feature hashgrid $\mathcal{H}$ and MLP weights $\mathcal{W}$), which gradually learns which regions to prioritize, such as a non-transparent object surface.
We split the space coarsely as $T\times T\times T$ uniform grids and the 3D tensor $\mathcal{G}$ stores trainable values representing the saliency of features near each grid voxel. Instead of letting multiple dense features in the same coarse grid share one value, we interpolate the values associated to the $8$ voxels surrounding the input coordinate $\bm{x}$ to get a smoothed saliency distribution $\mathcal{P}$ over the 3D space.
The trainable saliency grid represents in itself an opportunity to compress the NeRF model using differentiable model compression techniques, as discussed later in \S\ref{sec:admm}.

We initialize $\mathcal{G}$ as an all-ones tensor, where $T$ is typically much smaller than the resolution of the dense feature grid (In \S\ref{sec:setting}, we further investigate how different $T$ values affect the performance). 
Given a 3D coordinate $\bm{x}=(x, y, z)$, we fetch its corresponding feature vector $\bm{f}$ from the multi-level hashgrid~\cite{ngp}.
We then locate the $8$ voxels $(\bm{x}_1, \bm{x}_2, \ldots, \bm{x}_8)$ surrounding $\bm{x}$ in the 3D saliency grid $\mathcal{G}$, and fetch its corresponding saliency value $g_i$.
Specifically, the coordinate $\bm{x}_i = (x_i, y_i, z_i)$ are quantized to the values $(\hat{x}_i, \hat{y}_i, \hat{z}_i)$ between $0$ and $T-1$, which are then used to index the saliency grid $\mathcal{G}$.
This gives us the saliency value $g_i=\mathcal{G}(\hat{x}_i, \hat{y}_i, \hat{z}_i)$.
After obtaining $g_i$ for each of the $8$ saliency grid voxels surrounding $\bm{x}$, we calculate the \emph{saliency weight} $p$ at coordinate $\bm{x}$ by trilinear interpolation followed by a sigmoid operator, so that $0<p<1$.
The final weighted feature vector $\bm{v}$ is obtained by multiplying the saliency weight $p$ with the feature vector $\bm{f}$ fetched from the hashgrid, i.e., $\bm{v} = p\bm{f}$.

For mathematical analysis, we define the saliency distribution $\mathcal{P}$ as the collection of saliency weights $p$ at every possible coordinate $\bm{x}$, such that $p = \mathcal{P}(\bm{x})$. Our proposed method, \name, learns the saliency distribution $\mathcal{P}$ by training the saliency grid $\mathcal{G}$ using gradient descent.
It is worth noting that we inject $\mathcal{P}$ (and thus $\mathcal{G}$) to the training pipeline by feeding the weighted feature $\bm{v}$ instead of the raw feature $\bm{f}$ to the MLP for decoding;
the rationale for this choice is that
since the sigmoid function in the above workflow projects the saliency weights to the range $(0, 1)$, we can view the 3D saliency distribution as a probability distribution $\mathcal{P}(\bm{x})$ indicating the likelihood $p$ that a feature is non-zero at a given 3D position $\bm{x}$.
There are only two possibilities for the feature at a given position: either the position is visible with probability $p=\mathcal{P}(\bm{x})$, and we retrieve the feature vector $\bm{f}$ from the multi-level hashgrid as in~\cite{ngp}; otherwise the position is empty or occluded with probability $1-p$, and the corresponding feature vector is an all-zero vector $\bm{0}$.
Taking both cases into account, we can calculate the expectation $\bm{v}$ of the feature as:
\begin{align}
\bm{v} = \mathcal{P}(\bm{x}) \bm{f} + \left(1-\mathcal{P}(\bm{x})\right) \bm{0} = p \bm{f}\nonumber
\end{align}
Thus the weighted feature $\bm{v}$ fed to MLP decoder has a physical meaning, namely the expectation of the feature at $\bm{x}$ being either $\bm{f}$ fetched from feature hashgrid or $\bm{0}$.
Then training the saliency grid is equivalent to learning the probability distribution $\mathcal{P}(\bm{x})$.
This probability distribution represents the likelihood that a given voxel $\bm{x}$ contains a non-zero feature.
A feature with a lower saliency weight $p$ is more likely to be $\bm{0}$ and will cause less interference to other features assigned to the same bucket in the feature hashgrid.
When the saliency weight converges to $0$, that feature is pruned with a probability of $1$ and becomes a zero vector $\bm{0}$.

Training the saliency distribution can also be viewed as hash collision mitigation.
Instant-NGP employs hash encoding and distributes hash collisions evenly across voxels with non-zero density.
By learning to shuffle hash collisions across the space, \name guarantees less interference for more important features.
When the feature of a voxel is scaled by a higher saliency weight, it tends to accumulate more gradients during backward propagation, and the information at this voxel will dominate the trained feature value inside the hash bucket shared with other voxels.
On the other hand, a feature scaled by zero saliency weight has a zero gradient, hence causes no harm to colliding features.

\subsection{Soft zero-skipping gate}
\label{sec:gate}
\noindent While \name mitigates hash collisions by pruning unnecessary features to zero, the MLP used to decode these features into densities can break the sparsity in the 3D space: 
the MLP can be viewed as a function $\sigma=\mathcal{M}(\bm{v})$ mapping the feature vector $\bm{v}$ to the density $\sigma$, and typically $\mathcal{M}(\bm{0}) \neq 0$. 
Even though the saliency grid and ADMM pruner ensure most voxels direct to $\bm{0}$ features in the hashgrid, a typical MLP translates such a sparse feature domain to a non-sparse 3D space with non-zero densities scattered everywhere, which is evidently inaccurate. 
To enforce a $0$ output density when the input feature is $\bm{0}$, we introduce a zero-skipping gate $g(\bm{v})$ to the MLP decoder:
\begin{align}
&\mathcal{\hat{M}}(\bm{v}) = \hat{g}(\bm{v})\cdot \mathcal{M}(\bm{v})\nonumber\\
\mbox{where}\quad &\hat{g}(\bm{v}) := \tanh\left(\alpha \, \|\bm{v}\|_{2}\right)
\label{equ:soft_gate}
\end{align}

The gate $\hat{g}(\bm{v})$ is a differentiable approximation of the ``hard'' gate in Eq.~\eqref{equ:hard} to enforce the desired property $\mathcal{\hat{M}}(\bm{0}) = 0$, and we use a constant $\alpha$ to control the steepness of the 0-to-1 transition.
\begin{align}
\label{equ:hard}
{g}(\bm{v}):=\begin{cases}
			0, & \text{if}\quad \|\bm{v}\|_{2} = 0\\
            1, & \text{otherwise}
		 \end{cases}
\end{align}
The non-differentiable ``hard'' gate in Eq.~\eqref{equ:hard} may cause abrupt density value changes during training.
Also, the gradient $\frac{\partial g}{\partial p}$ of $g(\bm{v})=g(p\bm{f})$ is $0$ with $p \neq 0$, this $0$ gradient prevents the ADMM pruner from suppressing a small saliency value $p$ to 0, resulting in insufficient feature pruning and little improvement in hash collision.
Our experiments in \S\ref{sec:ablation} confirm that adding the ``hard'' gate $g(\bm{v})$ provides little gain over adding no gate.

Thus, we propose a ``soft'' zero-skipping gate $\hat{g}(\bm{v})$ in Eq.~\eqref{equ:soft_gate} that smoothly reduces the density output to zero for input features being pruned.
As an approximation of $g(\bm{v})$, the value of $\hat{g}(\bm{v})$ should smoothly transition from 0 to 1 for input $\bm{v}$ with small magnitudes $\|\bm{v}\|_2$, while otherwise staying close to 1 to avoid perturbing necessary information.
The $\tanh$ function is a suitable choice.
In our implementation we adjust the $\alpha$ value with a schedule, to gradually ``harden'' the soft gate as the training progresses.
Specifically, for epochs below $1000$, we set $\alpha$ to $10^4$ to ensure fast convergence.
Afterward we increase $\alpha$ to $10^5$ to minimize the perturbation to the fine-tuning process.

\subsection{ADMM pruner}
\label{sec:admm}

\noindent Training the 3D saliency grid improves the accuracy by redistributing hash collisions, which, However, cannot fully eliminate unnecessary features, as shown in Figure~\ref{fig:slice_no_pruner}. To address this, we introduce an ADMM pruner that enforces sparsity in the saliency grid $\mathcal{G}$ during training. 

We first motivate the use of ADMM pruner rigorously. Assuming a $K \times K \times K$ volume where each voxel $\bm{x}$ corresponds to a feature $\bm{f}$, and the ground truth feature $\bm{f}_1$ at voxel $\bm{x}_1 = (x_1, y_1, z_1)$ and $\bm{f}_2$ at a different voxel $\bm{x}_2 = (x_2, y_2, z_2)$ are typically different for a 3D scene.
Instant-NGP encodes this $K \times K \times K$ feature grid into a small hashgrid $\mathcal{H}$ whose size $<<(K \times K \times K)$. When $\bm{f}_1$ and $\bm{f}_2$ collide with each other by directing to the same hash bucket, their values queried from the hashtgrid become identical $\bm{f}_1^* = \bm{f}_2^* = \bm{f}_H$, where $\bm{f}_H$ is the trained feature stored in the shared hash bucket. Since the ground truth features $\bm{f}_1 \neq \bm{f}_2$ while the queried features $\bm{f}_1^* = \bm{f}_2^*$, the queried values differ from the ground truth, causing compression artifacts. 
To solve this problem, \name allows the colliding features to be different by introducing the scalar saliency weight $p$ to scale the queried features: $\bm{f}_1^* = p(\bm{x}_1) \bm{f}_H$ and $\bm{f}_2^* = p(\bm{x}_2) \bm{f}_H$. However, during training, the share feature $\bm{f}_H$ may not find a value that satisfies both equations, as they can be written as $\bm{f}_1^* = \frac{p(\bm{x}_1)}{p(\bm{x}_2)}\bm{f}_2^*$ which is not always true especially when $\bm{f}_1$ and $\bm{f}_2$ are non-zero. Then the queried features $\bm{f}_1^*$ and $\bm{f}_2^*$ may still cause compression artifacts (although less than instant-NGP). 
The ADMM pruner addresses this challenge by pruning the ground truth $\bm{f}_1$ or $\bm{f}_2$ to $\bm{0}$ when $\bm{x}_1$ or $\bm{x}_2$ is empty/invisible. For example, if voxel $\bm{x}_1$ is invisible from any view angle ($\bm{f}_1 = \bm{0}$), there exists a single $\bm{f}_H$ value to satisfy both $\bm{f}_1 = p(\bm{x}_1) \bm{f}_H$ and $\bm{f}_2 = p(\bm{x}_2) \bm{f}_H$, that is $\bm{f}_H = \bm{f}_2$, and there is no compression artifacts. The optimizer can converge to this optimal $\bm{f}_H$ value through gradient descent by learning that $p(\bm{x}_1)$ is $0$. 
In summary, making the feature grid sparse reduces the compression artifacts when encoding features into a hashgrid.

In what follows, we describe two methods for achieving sparsity: a basic L1-regularization approach and our ADMM pruner.
To sparsify the saliency grid $\mathcal{G}$, one simple approach is to add an L1 regularization term to the original MSE loss function $L$ typically used in NeRF training: 
\begin{align}
\label{equ:l1}
\min_{\mathcal{W}, \mathcal{H}, \mathcal{G}}  \left(L + \lambda  \|S_{\operatorname{sig}}\left(\mathcal{G}\right)\|_{1}\right)
\end{align}
where $\lambda$ is a constant weight to control the amount of sparsity and compression, and $\mathcal{H}$ and $\mathcal{W}$ denote respectively the embeddings in the multi-level hashgrid and the weights in the MLP. 
We use the sigmoid operator $S_{\operatorname{sig}}(\cdot)$ here because the forward pipeline (Figure~\ref{fig:design_all}) uses the saliency value after the sigmoid operator to weight the feature vector, and our ultimate goal is a sparse distribution of saliency value to prune unnecessary features. Note that $S_{\operatorname{sig}}(\mathcal{G})$ here denotes applying sigmoid operator on each element $g$ of the 3D tensor $\mathcal{G}$, and the result is a 3D tensor with all non-negative elements.
Therefore, the absolute operator in the L1 regularization term can be removed:
\begin{align}
\|S_{\operatorname{sig}}(\mathcal{G})\|_{1} = \sum_{g\in\mathcal{G}}|S_{\operatorname{sig}}(g)| = \sum_{g\in\mathcal{G}}S_{\operatorname{sig}}(g)\nonumber
\end{align}
Removing the absolute operator from the regularization term makes it differentiable and compatible with the training pipeline, eliminating the need to use more complex optimization algorithms such as proximal gradient descent to handle non-differentiable regularization terms.

Although the usage of L1 regularization is already a marked improved over the baseline (see \S\ref{sec:ablation}), we find it is often difficult to select a single $\lambda$ hyperparameter to control sparsity and compression.
During training when the 3D saliency grid $\mathcal{G}$ has already become sparse enough, the regularization term $\lambda \|S_{\operatorname{sig}}(\mathcal{G})\|_1$ may still provide gradients that further prune $\mathcal{G}$.
If $\lambda$ is set too high or if the algorithm is let to run for too long, visible features may be pruned resulting in performance degradation.

\begin{table*}[t]
\centering
\tabcolsep=0.11cm
\begin{tabular}{@{}cl|llllll@{}}
\toprule
\multicolumn{2}{l|}{\diagbox[dir=SE]{Method}{Param \#}{Hashgrid size}}                              &\multicolumn{1}{c}{$2^{14}$}       & \multicolumn{1}{c}{$2^{15}$}      & \multicolumn{1}{c}{$2^{16}$}      & \multicolumn{1}{c}{$2^{17}$}      & \multicolumn{1}{c}{$2^{18}$}      & \multicolumn{1}{c}{$2^{19}$}       \\ \midrule
\multicolumn{2}{l|}{Instant-NGP}                         & 0.50M  & 0.94M  & 1.77M & 3.34M & 6.22M & 11.49M \\ \midrule
\multicolumn{1}{c|}{\multirow{3}{*}{HollowNeRF}} & T=64  & 0.76M  & 1.21M & 2.04M & 3.60M & 6.48M & 11.75M \\
\multicolumn{1}{c|}{}                            & T=96  & 1.39M & 1.83M & 2.66M & 4.22M & 7.10M & 12.37M \\
\multicolumn{1}{c|}{}                            & T=128 & 2.60M & 3.04M & 3.87M & 5.43M & 8.32M & 13.58M \\ \bottomrule
\end{tabular}
\caption{The total number of model parameters, depending on the hashgrid size and saliency grid resolution $T$. This number counts the parameters in the hashgrid, MLP, and saliency grid.
(both \name and Instant-NGP employ a multi-resolution hashgrid with 16 levels. We use the maximum entries per level to denote the hashgrid size, following the notation in the Instant-NGP paper~\cite{ngp}. )
}
\label{table:cost}
\end{table*}

We use an ADMM pruner to overcome the limitations of simple L1 regularization. By introducing a configurable sparsity constraint $C$ such that $\|S_{\operatorname{sig}}(\mathcal{G})\|_1 < C$, feature pruning becomes the constrained optimization problem:
\begin{subequations}
\begin{align}
& \underset{\mathcal{M}, \mathcal{H}, \mathcal{G}}{\text{min}}
& &  L\left(\mathcal{M}, \mathcal{H}, \mathcal{G}\right) \nonumber \\
& \text{\quad s.t.}
& & \|S_{\operatorname{sig}}\left(\mathcal{G}\right)\|_1 < C \nonumber
\end{align}
\end{subequations}
We transform this constrained optimization problem into an equivalent unconstrained minimax problem in Problem~\eqref{equ:aug_lag} by using the augmented Lagrangian method which employs a trainable dual variable $\gamma$, also known as a Lagrange multiplier.
In essence, $\gamma$ serves as a trainable alternative to the fixed hyperparameter $\lambda$ in Problem~\eqref{equ:l1}.
\begin{align}
&\underset{\mathcal{W}, \mathcal{H}, \mathcal{G}}{\min}\   \underset{\gamma \ge 0}{\max} &&
 L\left(\mathcal{W}, \mathcal{H}, \mathcal{G}\right) + \frac{\rho_{\gamma}}{2} \left[\|S_{\operatorname{sig}}\left(\mathcal{G}\right)\|_1 - C \right]_{+}^2\nonumber\\&&& + \gamma \left(\|S_{\operatorname{sig}}\left(\mathcal{G}\right)\|_1 - C\right)
 \label{equ:aug_lag}
\end{align}
where $\rho_{\gamma}$ denotes the learning rate of dual variable $\gamma$, and $\left[\cdot\right]_{+}$ clamps the input to non-negative values.

At each iteration $t$, the optimizer first updates the coarse saliency grid $\mathcal{G}$, the feature hashgrid $\mathcal{H}$, and the MLP weights $\mathcal{W}$ to minimize the augmented loss function in Problem~\eqref{equ:aug_lag} using gradient descent, where the dual variable $\gamma$ is treated as a constant.
$\gamma$ is then updated using gradient ascent with $\gamma^{(t+1)} = \left[\gamma^{(t)} + \rho_{\gamma} \left(\|S_{\operatorname{sig}}\left(\mathcal{G}\right)\|_1 - C \right)\right]_{+}$

\section{Evaluation}
\label{sec:expe}

\noindent In this section, we first investigate the effect of different hyper-parameters, including hashgrid sizes, saliency grid sizes, and sparsity constraints, on the performance (\S\ref{sec:setting}).
We then conduct an ablation study to validate each of \name's design components (\S\ref{sec:ablation}).
Finally, we evaluate the performance of \name on various 3D scenes and compare it with the state-of-the-art works (\S\ref{sec:vsngp}). 

\subsection{Experimental setup}
\label{sec:impl}
\noindent We implement \name on top of \texttt{torch-ngp}~\cite{torch-ngp, torch-ngp1}, a PyTorch CUDA extension implementation of instant-NGP~\cite{ngp}.
To ensure a fair comparison, we evaluated both \name and instant-NGP using the same \texttt{torch-ngp} implementation.
Both methods use a 16-level hashgrid with a feature dimension of $2$, where the coarsest resolution of the feature grid is $16$, and the finest resolution is $1024$.
Both methods employ a MLP decoder that has only two hidden layers with a width of $64$ neurons, while \name augments the MLP with a zero-skipping gate as described in \S\ref{sec:gate}.

We use the peak signal-to-noise ratio (PSNR)   and the Learned Perceptual Image Patch Similarity (LPIPS)~\cite{lpips} with the default AlexNet backend as the performance metrics to assess the rendering quality.
Each experiments was conducted on a single NVIDIA Tesla A100 GPU for both training and evaluation.
The machine used was equipped with an AMD EPYC 7742 64-Core CPU, 1TB of memory, and 8 GPU cards in total.

Unless otherwise specified, we train the model for $300,000$ steps using the Adam optimizer with an initial learning rate of $1 \times 10^{-2}$. Further details regarding the appropriate selection of \name hyper-parameters, such as $T$ and $C$, are discussed in \S\ref{sec:setting}.

\subsection{Investigating the cost-accuracy tradeoff}
\label{sec:setting}
\begin{figure}[t]
\begin{subfigure}[b]{0.99\linewidth}
    \centering
    \includegraphics[width=\linewidth]{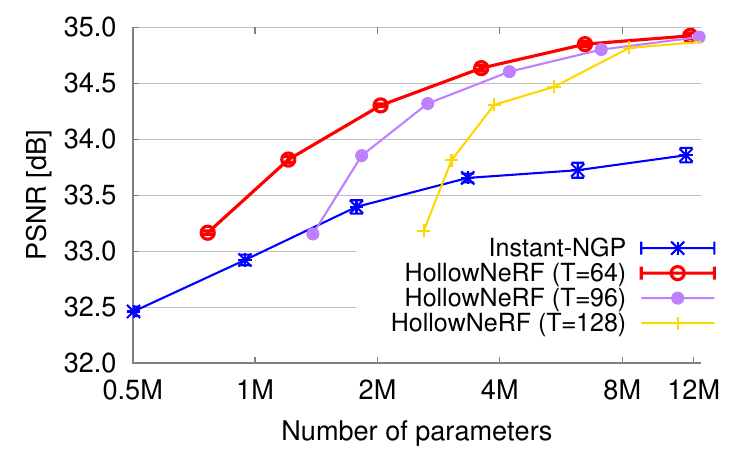}
    \caption{PSNR $\uparrow$ (higher is better).}
 \label{fig:t_psnr}
  \end{subfigure}\\
  \begin{subfigure}[b]{0.99\linewidth}
  \centering
    \includegraphics[width=\linewidth]{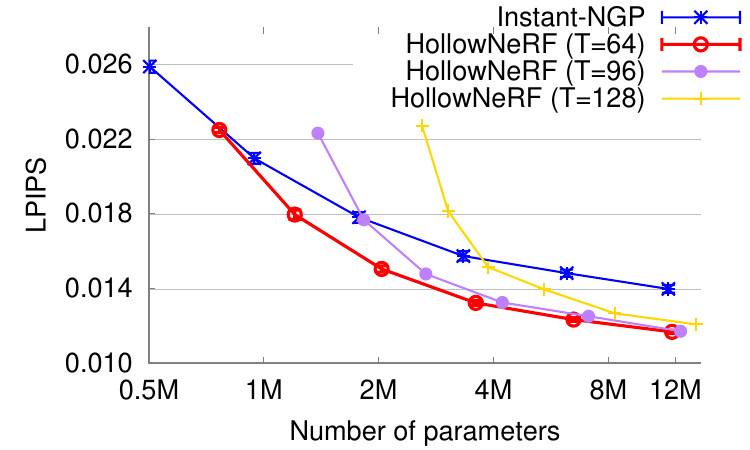}
    \caption{LPIPS $\downarrow$ (lower is better).}
 \label{fig:t_lpips}
  \end{subfigure}
   \caption{Performance vs. saliency grid size $T$.}
\label{fig:t_all}
\end{figure}

\begin{figure}[t]
\begin{subfigure}[b]{0.99\linewidth}
    \centering
    \includegraphics[width=\linewidth]{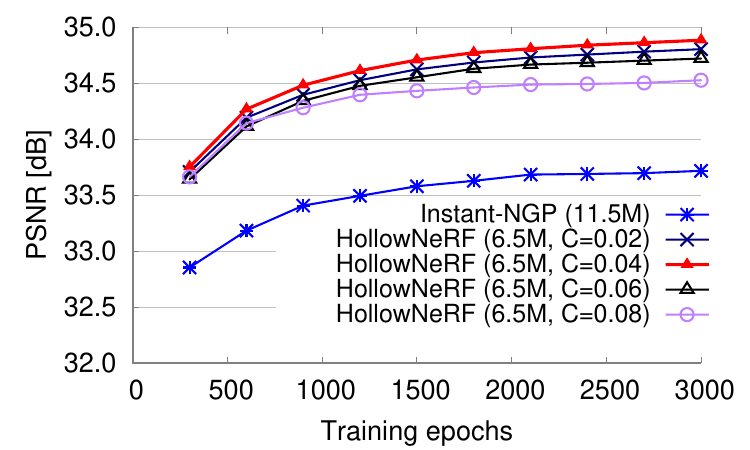}
    \caption{PSNR $\uparrow$ (higher is better).}
 \label{fig:c_psnr}
  \end{subfigure}\\
  \begin{subfigure}[b]{0.99\linewidth}
  \centering
    \includegraphics[width=\linewidth]{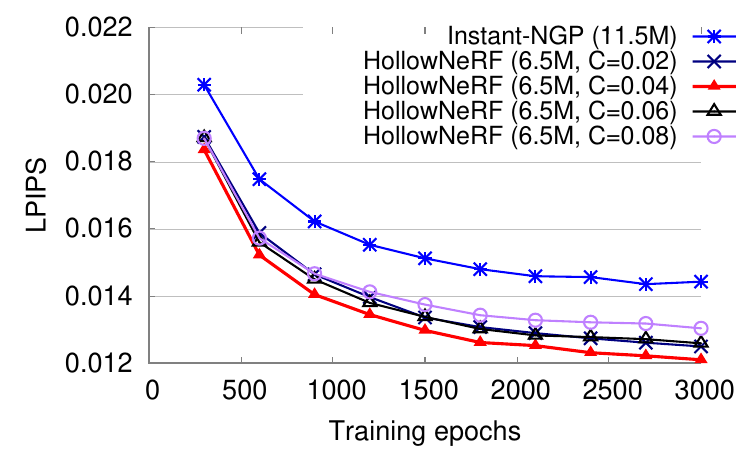}
    \caption{LPIPS $\downarrow$ (lower is better).}
 \label{fig:c_lpips}
  \end{subfigure}
   \caption{Performance vs. sparsity constraint $C$.}
\label{fig:c_all}
\end{figure}

\begin{figure}[t]
\begin{subfigure}[b]{0.99\linewidth}
    \centering
    \includegraphics[width=\linewidth]{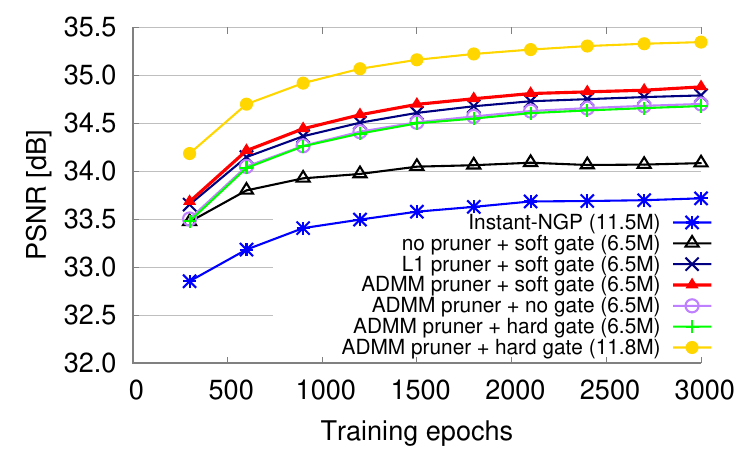}
    \caption{PSNR $\uparrow$ (higher is better).}
 \label{fig:abl_psnr}
  \end{subfigure}\\
  \begin{subfigure}[b]{0.99\linewidth}
  \centering
    \includegraphics[width=\linewidth]{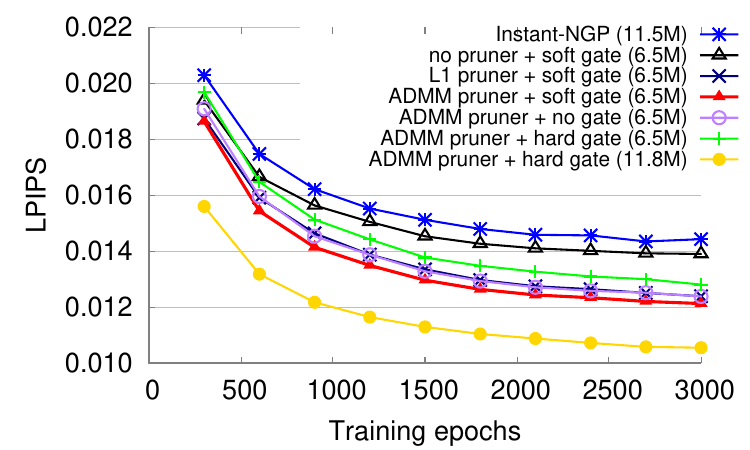}
    \caption{LPIPS $\downarrow$ (lower is better).}
 \label{fig:abl_lpips}
  \end{subfigure}
   \caption{Ablation study for the design components.}
\label{fig:abl_all}
\end{figure}

\begin{figure*}
  \adjustbox{minipage=7em,valign=c}{\subcaption{\name, 6.48M}\label{fig:our_18}}
  \begin{subfigure}[t]{0.21\linewidth}
  \centering
  \stackon[3pt]{\includegraphics[width=.99\linewidth,valign=c]{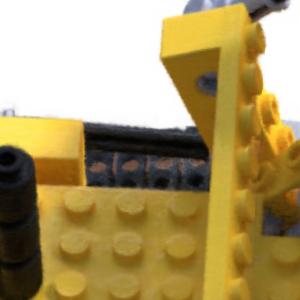}}{Lego}
  \end{subfigure}%
  \begin{subfigure}[t]{0.21\linewidth}
  \centering
  \stackon[3pt]{\includegraphics[width=.99\linewidth,valign=c]{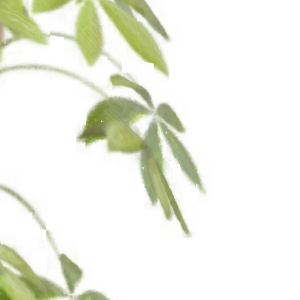}}{Ficus}
  \end{subfigure}%
  \begin{subfigure}[t]{0.21\linewidth}
  \centering
  \stackon[3pt]{\includegraphics[width=.99\linewidth,valign=c]{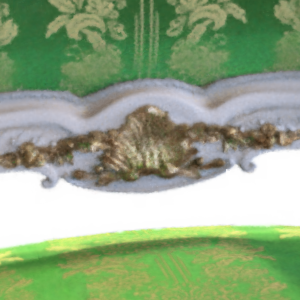}}{Chair}
  \end{subfigure}%
  \begin{subfigure}[t]{0.21\linewidth}
  \centering
  \stackon[3pt]{\includegraphics[width=.99\linewidth,valign=c]{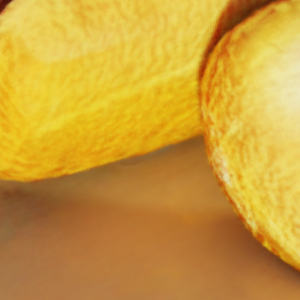}}{Hotdog}
  \end{subfigure}%
  \\
  \adjustbox{minipage=7em,valign=c}{\subcaption{\name, 11.75M}\label{fig:our_19}}%
  \begin{subfigure}[t]{0.21\linewidth}
  \centering
  \includegraphics[width=.99\linewidth,valign=c]{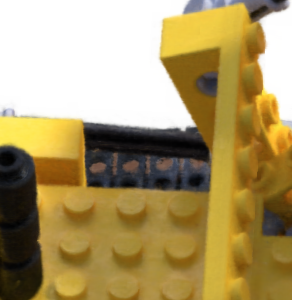}
  \end{subfigure}%
  \begin{subfigure}[t]{0.21\linewidth}
  \centering
  \includegraphics[width=.99\linewidth,valign=c]{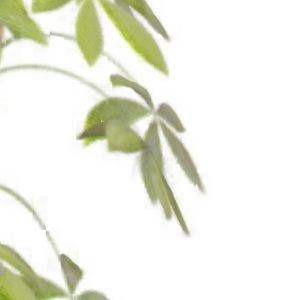}
  \end{subfigure}%
  \begin{subfigure}[t]{0.21\linewidth}
  \centering
  \includegraphics[width=.99\linewidth,valign=c]{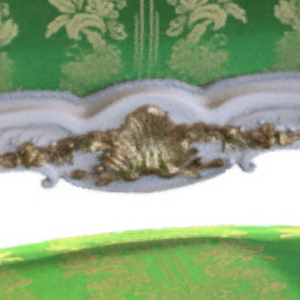}
  \end{subfigure}%
  \begin{subfigure}[t]{0.21\linewidth}
  \centering
  \includegraphics[width=.99\linewidth,valign=c]{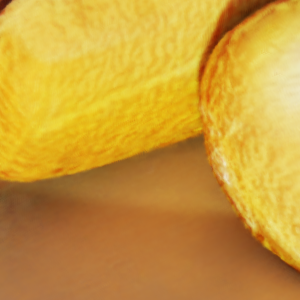}
  \end{subfigure}%
  \\
  \adjustbox{minipage=7em,valign=c}{\subcaption{Instant-NGP, 11.49M}\label{fig:ngp_19}}%
  \begin{subfigure}[t]{0.21\linewidth}
  \centering
  \includegraphics[width=.99\linewidth,valign=c]{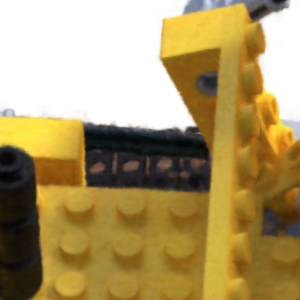}
  \end{subfigure}
  \begin{subfigure}[t]{0.21\linewidth}
  \centering
  \includegraphics[width=.99\linewidth,valign=c]{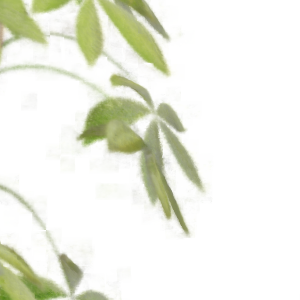}
  \end{subfigure}
  \begin{subfigure}[t]{0.21\linewidth}
  \centering
  \includegraphics[width=.99\linewidth,valign=c]{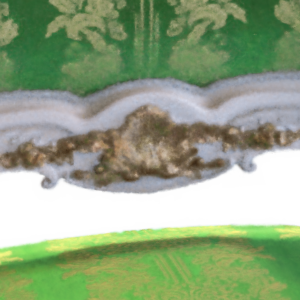}
  \end{subfigure}
  \begin{subfigure}[t]{0.21\linewidth}
  \centering
  \includegraphics[width=.99\linewidth,valign=c]{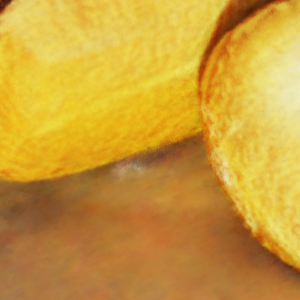}
  \end{subfigure}
  \caption{Qualitative comparison with instant-NGP over scenes from the NeRF synthetic dataset.}
  \label{fig:vis_all}
\end{figure*}

\begin{figure*}[t]
\begin{center}
\includegraphics[width=0.96\linewidth]{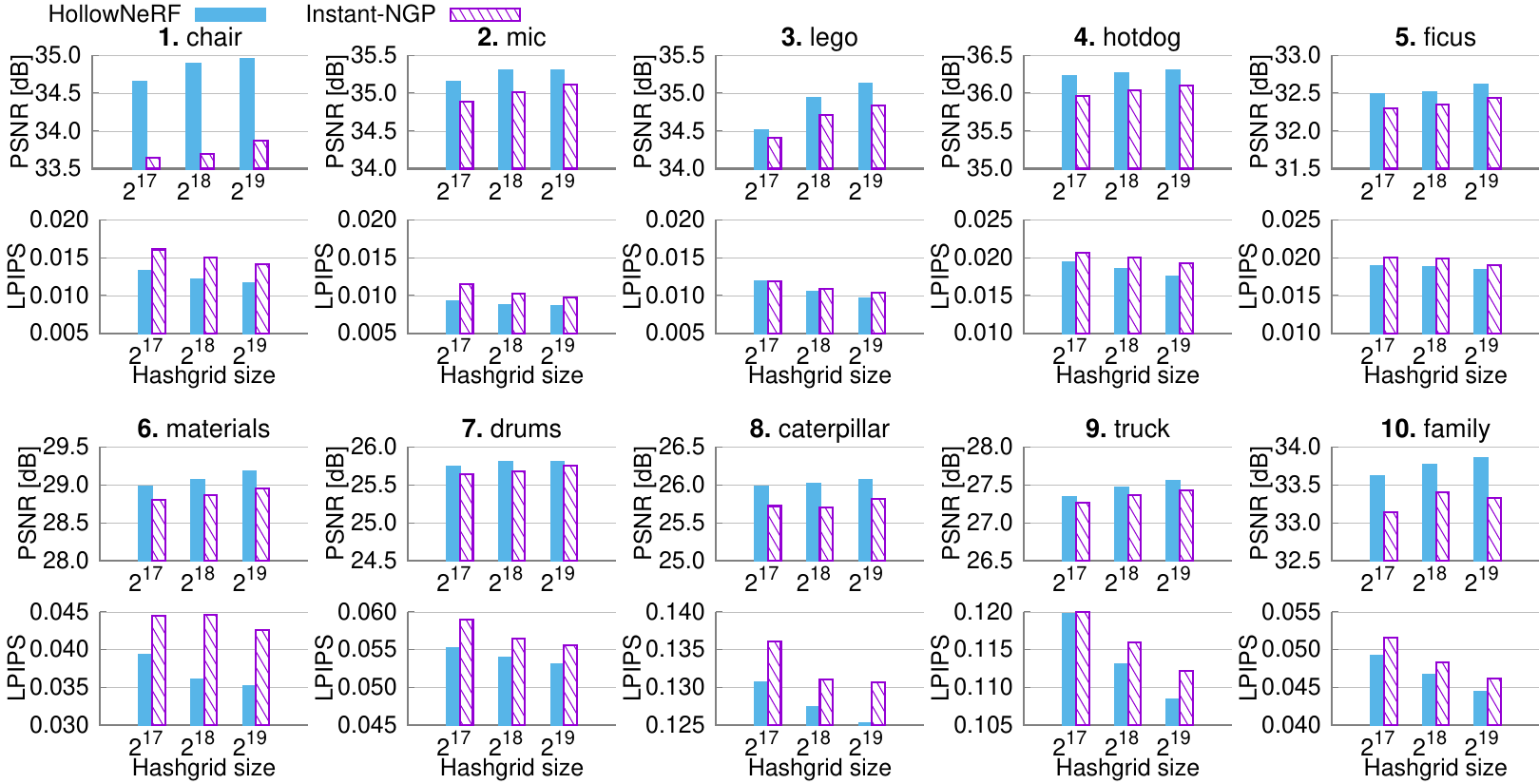}
\end{center}
   \vspace{-1em}
   \caption{Comparison of PSNR($\uparrow$) and LPIPS ($\downarrow$) performance between \name and Instant-NGP on various scenes. }
\label{fig:comp_sets}
\end{figure*}

\begin{figure}[t]
\begin{center}
\resizebox{0.99\linewidth}{!}
{\input{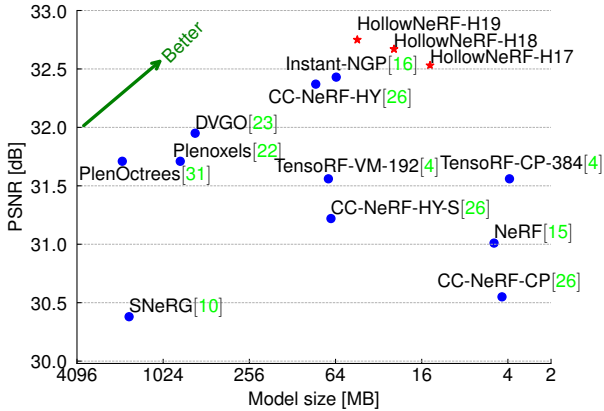}}
\end{center}
   \vspace{-1em}
   \caption{Comparison of \name and recent methods in terms of the trade-off between PSNR and model size (``\name-H19'' means using a hashgrid size of $2^{19}$).}
\label{fig:comp_scatter}
\end{figure}

\noindent We first investigate the cost-accuracy tradeoff of \name to determine the appropriate hyper-parameter configuration for larger scale experiments in \S\ref{sec:vsngp}.
We conducted experiments on the chair scene from the NeRF synthetic dataset, varying the hashgrid size from $2^{14}$ to $2^{19}$ and testing the saliency grid resolution of $64$, $96$, and $128$ for each hashgrid size. 
Table~\ref{table:cost} summarizes the cost of different configurations, including the number of parameters in the hashgrid, MLP, and saliency grid. 
We use a sparsity constraint of $C=0.04$ in this experiment. 

The results in Figure~\ref{fig:t_all} indicate that increasing the saliency grid resolution $T$ from $64$ to $96$ or $128$ does not improve accuracy much but results in a higher parameter count, which harms the cost-accuracy tradeoff. On the other hand, lowering $T$ below $64$ causes unstable convergence. Consequently, we use a grid resolution of $T=64$ for the subsequent experiments.
Figure~\ref{fig:t_all} further shows that \name outperforms the baseline over a wide range of model sizes, achieving higher PSNR and lower LPIPS values for comparable parameter counts.
The best overall model, \name with $T=64$, uses $6.48$M parameters to attain a $34.85$dB PSNR, as compared to Instant-NGP's $33.86$dB using $11.49$M.
In other words, it achieves about $1$dB higher PSNR than Instant-NGP, while utilizing only $56\%$ of the parameters employed by Instant-NGP.

We also repeated the training and evaluation for each hashgrid size five times for \name with $T=64$ and instant-NGP, and plotted the standard deviations across the five tests as error bars in Figure~\ref{fig:t_all}.
The results demonstrate that our approach is more stable than Instant-NGP in terms of convergence, with lower performance variation.

To investigate the impact of the choice of sparsity constraint $C$ on performance, we conducted additional experiments following the same setup as previous tests, while varying the $C$ values, and report the results in Figure~\ref{fig:c_all}. First, from Figure~\ref{fig:c_psnr}, we observe that the $C=0.04$ configuration yields the highest PSNR, in other words, \name achieves the highest accuracy when $4\%$ of the spatial voxels are non-zero. $C=0.02$ exhibits slightly diminished accuracy, potentially due to excessive pruning in the 3D space, which compromises crucial information for scene reconstruction. Conversely, $C=0.08$ shows the lowest accuracy across all tested \name configurations, suggesting that keeping $8\%$ non-zero voxels causes excessive hash collisions when packing them into the hashgrid, and the 3D scene to capture could be sparser. We have similar observation for LPIPS, as depicted in Figure~\ref{fig:c_lpips}.

\subsection{Ablation study}
\label{sec:ablation}
\noindent In this ablation study, we analyze the impact of each component in our solution by comparing the convergence speed and resulting performance of the configurations with and without each design component.
We use a saliency grid resolution of $64$ and a hashgrid size of $2^{18}$ (total $6.48$M parameters) for all \name configurations, while the baseline is instant-NGP with a hashgrid size of $2^{19}$ (total $11.49$M parameters).
To investigate how each component affects the convergence speed, we evaluate the performance on the test sets every $300$ epochs during the total $3000$ epochs.

The results presented in Figure~\ref{fig:abl_all} show that each design component presence gracefully increases the overall performance, without any catastrophic failure.
Each of the \name ablated configurations still outperforms instant-NGP, while employing less parameters.

\subsection{Comparisons on standardized datasets}
\label{sec:vsngp}
\noindent Finally, we evaluate the performance of \name on 3D scenes from the NeRF synthetic dataset~\cite{nerf} and the more complicated Tanks and Temples dataset~\cite{tanks}.
Based on the cost-accuracy tradeoff investigated in Sec.~\ref{sec:setting}, we set \name's hyper-parameters to a sparsity constraint of $C=0.04$ and a saliency grid resolution of $T=64$.

We first provide qualitative example views comparing \name with Instant-NGP. The distinct advantages of \name are evident through the visuals showcased in Figure~\ref{fig:vis_all}. For example, from the lego scene in column 1, we observe that \name with $6.48$M parameters achieves notably superior rendering accuracy over Instant-NGP with $77\%$ more parameters ($11.49$M). The advantage is particularly pronounced within the bulldozer track region. Furthermore, \name with $11.75$M parameters exhibits the highest accuracy, capturing fine details such as the tiny indentation on the edge of the black rod located at the left border of the view.

Then we show the PSNR and LPIPS performance of \name and Instant-NGP on various 3D scenes (we sample 10 different scenes due to the page limit) in Figure~\ref{fig:comp_sets}. 
For each scene, we test three hashgrid sizes ($2^{17}$, $2^{18}$, $2^{19}$) for both methods.
We can observe that \name consistently outperforms Instant-NGP on all the 10 scenes tested, achieving better PSNR and LPIPS performance with fewer parameters when using the same hashgrid size.
For seven out of ten tested scenes, the \name with the smallest hashgrid size ($2^{17}$, for a total parameter count including MLP of $3.34$M) achieves higher PSNR than the best Instant-NGP model (hashgrid size of $2^{19}$, corresponding to $11.49$M parameters including the MLP).

Finally, we compare the average PSNR across the NeRF synthetic dataset~\cite{ngp} and the corresponding model size for \name and several recent works. Figure~\ref{fig:comp_scatter} shows that \name achieves the best cost-accuracy tradeoff among all tested methods. In particular, \name with a hashgrid size of $2^{17}$ achieves an average PSNR of $32.53$dB with only a mode size of $14.0$MB.

\section{Discussion}
\label{sec:limitations}
\noindent The compression gains of \name require the 3D scene being sparse, \emph{i.e.} mostly filled with either empty or occluded space, which holds for typical 3D scenes.
\name can still operate when this assumption does not hold, such as for scenes containing smoke, fire, clouds, but its performance will regress to the baseline without pruning.

The current version of \name, like its predecessor Instant-NGP, faces challenges in modeling objects with reflective surfaces. In future works, we plan to extend the concept of \name to the  frameworks~\cite{reflective, netrf, nerfosr, refnerf, nerfinthedark} with better support for reflective surfaces.

\section{Conclusions}
\label{sec:conclusion}
\noindent In this paper, we present \name, a novel hashgrid-based NeRF technique that achieves superior rendering quality than state-of-the-art solutions like instant-NGP while using only a fraction of the parameters. \name mitigates hash collisions by a simple feature pruning mechanism. Unlike existing methods that rely on explicit surface geometries for feature pruning, \name learns a 3D saliency grid to guide feature compression during training, and employs an ADMM pruner to enforce a sparse feature domain.
Our experiments demonstrate that \name achieves a better balance between cost and accuracy than state-of-the-art solutions, making a solid step towards lightweight and ubiquitous NeRF applications.

{\small
\bibliographystyle{ieee_fullname}
\bibliography{main}
}

\end{document}